\documentclass{article} % For LaTeX2e
\usepackage{iclr2027_conference,times}

\usepackage{amsmath,amsfonts,bm}

\def\eqref#1{equation~\ref{#1}}
\def\1{\bm{1}}

\DeclareMathAlphabet{\mathsfit}{\encodingdefault}{\sfdefault}{m}{sl}
\SetMathAlphabet{\mathsfit}{bold}{\encodingdefault}{\sfdefault}{bx}{n}

\usepackage{microtype}
\usepackage{graphicx}
\usepackage{subcaption}
\usepackage{booktabs} % for professional tables
\usepackage{tabularx}

\usepackage[hidelinks]{hyperref}
\usepackage{amsmath}
\usepackage{amssymb}
\usepackage{mathtools}
\usepackage{amsthm}
\usepackage{multirow}

\usepackage{url}
\usepackage{float}
\usepackage{microtype}
\usepackage{graphicx,wrapfig,lipsum}
\usepackage{subcaption}
\usepackage{booktabs} % for professional tables
\usepackage{colortbl}
\usepackage{xspace}
\usepackage{xcolor}
\definecolor{softgreen}{RGB}{90,150,120}
\title{Video Prediction Policy 2: Predict Better, \\Act Better}

\author{%
\begin{minipage}[t]{\dimexpr\textwidth-2\tabcolsep\relax}
\raggedright\normalfont\mdseries
\resizebox{\linewidth}{!}{%
\begin{tabular}{@{}l@{}}
\mbox{\textbf{Yanjiang Guo}\textsuperscript{\mdseries 1,2,*}, \textbf{Haodong Yan}\textsuperscript{\mdseries 1,3,*},
\textbf{Zhide Zhong}\textsuperscript{\mdseries 1,3,*}, \textbf{Zhongru Zhang}\textsuperscript{\mdseries 2,*},
\textbf{Qingyuan Yang}\textsuperscript{\mdseries 1,2,*}}\\
\mbox{\textbf{Qingzhou Lu}\textsuperscript{\mdseries 1,2}, \textbf{Xiaoyu Chen}\textsuperscript{\mdseries 2},
\textbf{Yen-Jen Wang}\textsuperscript{\mdseries 4}, \textbf{Shuying Deng}\textsuperscript{\mdseries 1,2,4},
\textbf{Chenghan Yang}\textsuperscript{\mdseries 2}, \textbf{Puzhen Yuan}\textsuperscript{\mdseries 1,2}}\\
\mbox{\textbf{Chenxin Liu}\textsuperscript{\mdseries 1,2}, \textbf{Tun Ban}\textsuperscript{\mdseries 1,5},
\textbf{Xiang Zhu}\textsuperscript{\mdseries 1,2}, \textbf{Yichen Liu}\textsuperscript{\mdseries 1,2},
\textbf{Kun Feng}\textsuperscript{\mdseries 1,2}, \textbf{Haoang Li}\textsuperscript{\mdseries 3},
\textbf{Jianyu Chen}\textsuperscript{\mdseries 1,2}}
\end{tabular}%
}\\[0.5em]
\normalfont\small
\textsuperscript{*}Equal Contribution\quad
\textsuperscript{1}Robotera\quad \textsuperscript{2}Tsinghua University\quad
\textsuperscript{3}HKUST (GZ) \\
\textsuperscript{4}University of California, Berkeley\quad
\textsuperscript{5}Shanghai Jiaotong University\\
Project Page: \url{https://robert-gyj.github.io/video-prediction-policy-2}\\
Code Page: \url{https://github.com/roboterax/video-prediction-policy-2}
\end{minipage}%
}

\iclrfinalcopy % Uncomment for camera-ready version, but NOT for submission.
\begin{document}

\maketitle
\vspace{-3mm}
\begin{abstract}
% World action models (WAMs) have emerged as an important class of generalist robot policies, aiming to translate the predictive capabilities of video foundation models into generalizable robot control. However, our evaluations reveal that incorporating action learning often substantially degrades video prediction generalization. On unseen, open-ended manipulation tasks, existing WAMs frequently produce incorrect motion predictions that lead to erroneous actions during execution. 

% We first train a strong video model for manipulation. Our central finding is that reducing uncertainty in future prediction is key to generalization across manipulation tasks. Guided by this insight, we annotate video clips with detailed captions and perform event-level video pretraining. 

World action models (WAMs) have emerged as an important class of generalist robot policies, aiming to transfer video prediction priors to action learning. However, we find that existing WAMs frequently produce incorrect motion predictions in open-ended environments, leading to erroneous actions. We attribute this limitation to two factors: (1) base video models are not optimized for manipulation, and (2) naively incorporating action components into video models can substantially degrade their generalization capabilities. We introduce Video Prediction Policy 2 (VPP2), a WAM that enables strong zero-shot generalization in both video prediction and action generation. First, we curate a large-scale, diverse dataset of manipulation videos to continue pretraining the base video foundation model. We annotate video clips with detailed captions and perform \textit{event-level} video pretraining to promote generalization across open-ended manipulation tasks. Second, we post-train and distill the video model into a single-step visual planner with a fixed prediction horizon. Finally, we introduce an action module via a mixture-of-transformers (MoT) architecture to learn an implicit inverse dynamics model. Experiments demonstrate three key results: (1) VPP2-14B outperforms Cosmos3-64B by 11.0\% points in video prediction instruction-following success rate on open-ended tasks; (2) VPP2 surpasses the strongest baseline by 18.5\% points in success rate on real-world zero-shot ALOHA manipulation tasks; and (3) following benchmark-specific post-training, VPP2 achieves the highest success rates among evaluated methods on the challenging LIBERO-Pro, LIBERO-OOD, and RoboDojo benchmarks.

% The abstract paragraph should be indented 1/2~inch (3~picas) on both left and
% right-hand margins. Use 10~point type, with a vertical spacing of 11~points.
% The word \textsc{Abstract} must be centered, in small caps, and in point size 12. Two
% line spaces precede the abstract. The abstract must be limited to one
% paragraph.
% The abstract paragraph should be indented 1/2~inch (3~picas) on both left and
% right-hand margins. Use 10~point type, with a vertical spacing of 11~points.
% The word \textsc{Abstract} must be centered, in small caps, and in point size 12. Two
% line spaces precede the abstract. The abstract must be limited to one
% paragraph.
% The abstract paragraph should be indented 1/2~inch (3~picas) on both left and
% right-hand margins. Use 10~point type, with a vertical spacing of 11~points.
% The word \textsc{Abstract} must be centered, in small caps, and in point size 12. Two
% line spaces precede the abstract. The abstract must be limited to one
% paragraph.
% The abstract paragraph should be indented 1/2~inch (3~picas) on both left and
% right-hand margins. Use 10~point type, with a vertical spacing of 11~points.
% The word \textsc{Abstract} must be centered, in small caps, and in point size 12. Two
% line spaces precede the abstract. The abstract must be limited to one
% paragraph.
\end{abstract}

\begin{figure}[h]
\vspace{-3mm}
    \centering
    \includegraphics[width=1.0\linewidth]{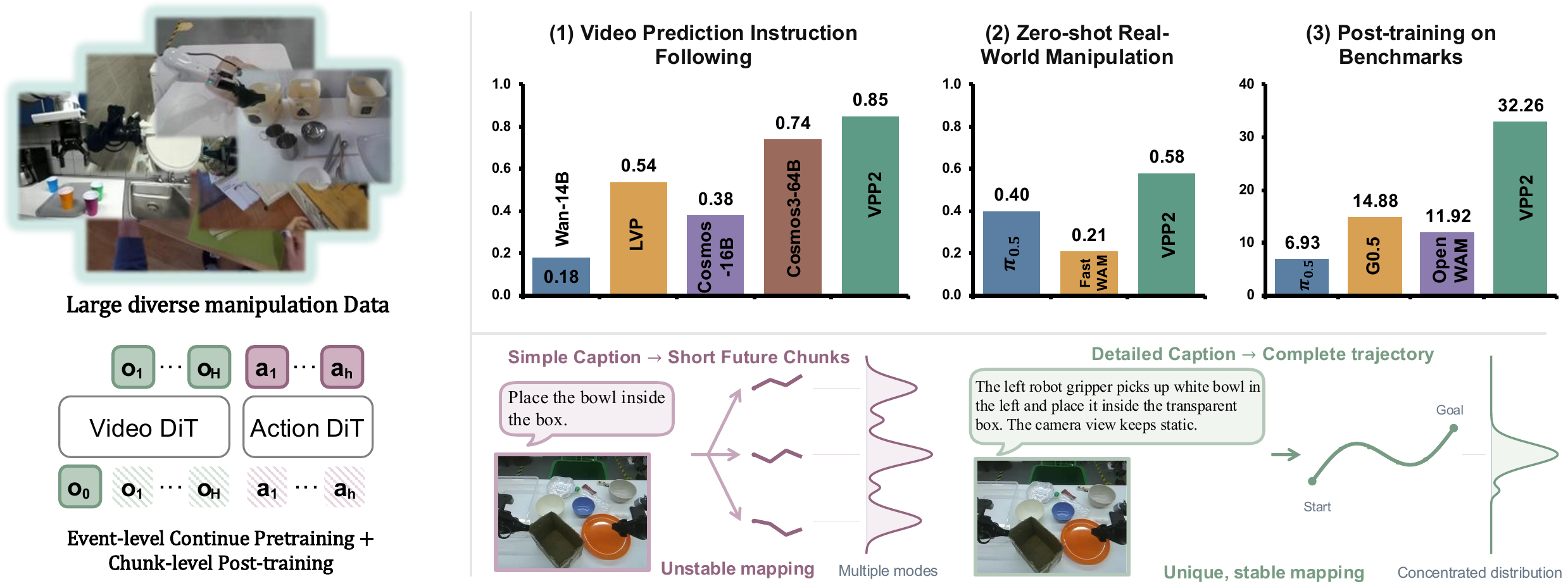}
    \caption{A typical robot policy maps a simple instruction and an observation into a short action chunk. As datasets scale, this mapping becomes increasingly multimodal and uncertain, leading models to learn spurious short-horizon correlations. VPP2 first establishes consistent semantic-to-trajectory mappings via \textit{next event video prediction} with detailed captions, then post-trains and distills the model to generate short action chunks.}
    \label{fig:fig1}
    \vspace{-5mm}
\end{figure}

\section{Introduction}
\vspace{-2mm}
% World action models (WAMs) have emerged as an important class of generalist robot policies~\citep{hu2024video,liao2025genie,kim2026cosmospolicy,yuan2026fastwam,ma2026dit4dit}. 
% By building on video foundation models pretrained on internet-scale data~\citep{wan2025wan,agarwal2025cosmos}, WAMs use video prediction priors to guide action learning, with the promise of transferring the rich physical and semantic knowledge of video models into generalizable robot control.

World action models (WAMs) are developing rapidly and have become an important class of generalist robot policies~\citep{hu2024video,liao2025genie,kim2026cosmospolicy,yuan2026fastwam,ma2026dit4dit,li2026lingbotva,li2026wallwm,agibot2026geact2,ye2026dreamzero}. Many WAMs build on video foundation models~\citep{wan2025wan,agarwal2025cosmos,yang2025cogvideox}, aiming to transfer their rich priors about physical dynamics to policy learning. This premise is compelling: accurate, instruction-conditioned video predictions can guide action generation through an explicit or implicit inverse dynamics model~\citep{du2024learning,hu2024video}. However, prior work~\citep{zhang2026qwen} has found that existing WAMs often perform well only on a narrow set of seen tasks, while producing unreliable future motion predictions in unseen, out-of-distribution scenarios. These prediction failures, in turn, lead to erroneous actions. In recent benchmarks that require generalization, robot policies, including WAMs, with near-perfect success rates on standard evaluation suites suffer substantial performance drops under perturbed task configurations and novel skill compositions~\citep{zhou2025liberopro,li2025vlas,mishra2026understanding}.

% We identify two key factors that lead to the weakness of WAMs: First, base video models are optimized for general-purpose content creation rather than manipulation, and therefore often fail to follow manipulation instructions faithfully. Second, after WAMs introduce additional action components or leaning target to pretrained-video foundation models, the generalization of the video models are largely demaged.

We identify two key factors that limit the generalization of existing WAMs to unseen scenarios. First, base video models are primarily optimized for creative content generation and aesthetic quality rather than precise physical dynamics, and therefore often fail to follow manipulation instructions faithfully~\citep{chen2025large,zhang2026qwen}. Second, introducing action-specific components or training objectives into pretrained video foundation models can substantially degrade their generalization capabilities~\citep{mishra2026understanding}.

To address these issues, we introduce Video Prediction Policy 2 (VPP2), a WAM that achieves strong zero-shot generalization in both video prediction and action generation. Our first objective is to train a generalizable video foundation that can faithfully follow instructions and make future predictions grounded in the current observation, without unsupported changes to the scene or task. To this end, we annotate each clip with a detailed caption specifying the manipulation process, active end effector, and target object, and perform \textit{event-level} video prediction training. As illustrated in Figure~\ref{fig:fig1}, detailed captions and complete event-level trajectory prediction substantially reduce uncertainty in future prediction and encourage consistent mappings from semantic descriptions to visual trajectories, improving generalization across diverse instructions. We then post-train the video model to predict fixed-horizon future chunks and apply consistency distillation~\citep{song2023consistency} to obtain a single-step visual planner for real-time robot execution. Finally, we introduce an action expert through a mixture-of-transformers (MoT) architecture~\citep{liang2025mixture} to learn an implicit inverse dynamics model conditioned on the predicted future. During policy execution, we pair VPP2 with a VLM planner~\citep{bai2025qwen3,seed2026seed2} that translates open-ended instructions into explicit subtask instructions.

Our experiments demonstrate three key advantages of VPP2. (1) VPP2-14B follows manipulation instructions faithfully, outperforming Cosmos3-64B~\citep{agarwal2026cosmos} by 11\% in instruction-following success rate on open-ended tasks. (2) VPP2 exhibits strong zero-shot manipulation capabilities on a real-world ALOHA platform~\citep{zhao2023learning}, achieving an average success rate of 58.5\% across 10 task categories, compared with 40.0\% for $\pi_{0.5}$~\citep{intelligence2025pi_} and 20.5\% for Fast-WAM~\citep{yuan2026fastwam}. (3) Following benchmark-specific post-training, VPP2 achieves the highest success rates among evaluated methods on LIBERO-Pro~\citep{zhou2025liberopro} (45.0\%, compared with 11.0\% for the strongest baseline), LIBERO-OOD~\citep{li2025vlas,mishra2026understanding} (63.9\%), and RoboDojo~\citep{chen2026robodojo} (32.26\%, state-of-the-art).

\vspace{-1mm}
\section{Data Processing Pipeline}
\vspace{-1mm}
In this section, we describe our data collection and curation pipeline, including how we unify the inputs and outputs across diverse manipulation datasets.

\textbf{Video Sources.}
We compile a large-scale, diverse collection of manipulation videos for continued pretraining of an open-source video foundation model~\citep{wan2025wan}. Our data span three categories: robot manipulation datasets, human activity datasets, and general-purpose video datasets. 
For robot manipulation, we incorporate nearly all publicly available, high-quality datasets that meet our selection criteria. We also additionally collected many robot and human ego-centric datasets by ourselves.
For general-purpose video datasets, we apply an extensive filtering pipeline similar to that of LVP~\citep{chen2025large} to retain only videos depicting manipulation-related activities. 

% Table~\ref{table:video_source} summarizes the datasets used for training; full dataset references are provided in Appendix~\ref{app_data}.

% \begin{table}[h]
%     \centering
%     \small
%     \setlength{\tabcolsep}{4pt}
%     \renewcommand{\arraystretch}{1}
%     \begin{tabularx}{\linewidth}{@{}llX@{}}
%         \toprule
%         Data Type & Embodiment Type & Data Sources \\
%         \midrule
%         \multirow{3}{*}{Robot}
%         & Single-arm
%         & OXE, RoboMIND, DROID, RH20T, MolmoAct \\
%         & Dual-arm
%         & AgiBot World-Beta, RoboCOIN, RDT, GM-100, ABC-130K, Self-collected ALOHA Data\\
%         & Mobile \& humanoid
%         & InternData-A1, Galaxea Open-World, Self-collected dexterous hands \\
%         \midrule
%         \multirow{2}{*}{Human}
%         & \multirow{2}{*}{Human hands}
%         & Open-source: EgoDex, Ego4D, EPIC-KITCHENS, Something-Something V2, VITRA-1M \\
%         & & Self-collected egocentric data\\
%         \midrule
%         General Video
%         & Human hands
%         & OpenVid, Panda-70M \\
%         \bottomrule
%     \end{tabularx}
%     \caption{Overview of training data sources. Full dataset references are provided in Appendix~\ref{app_data}, Table~\ref{table:video_source_full}.}
%     \label{table:video_source}
%     \vspace{-3mm}
% \end{table}

\textbf{Video Filtering, Segmentation.}
We first remove corrupted trajectories, including those with camera failures or abrupt discontinuities, and segment the remaining trajectories into semantically coherent clips lasting 1--12 seconds. We identify candidate boundaries using heuristic cues associated with natural transitions, such as local minima in human-hand or end-effector velocity and moments when the gripper or fingers open or close. A vision-language model (VLM) then refines the segmentation by merging adjacent clips where appropriate and adjusting their start and end points.

\begin{figure}[t]
    \centering
    \includegraphics[width=1.0\linewidth]{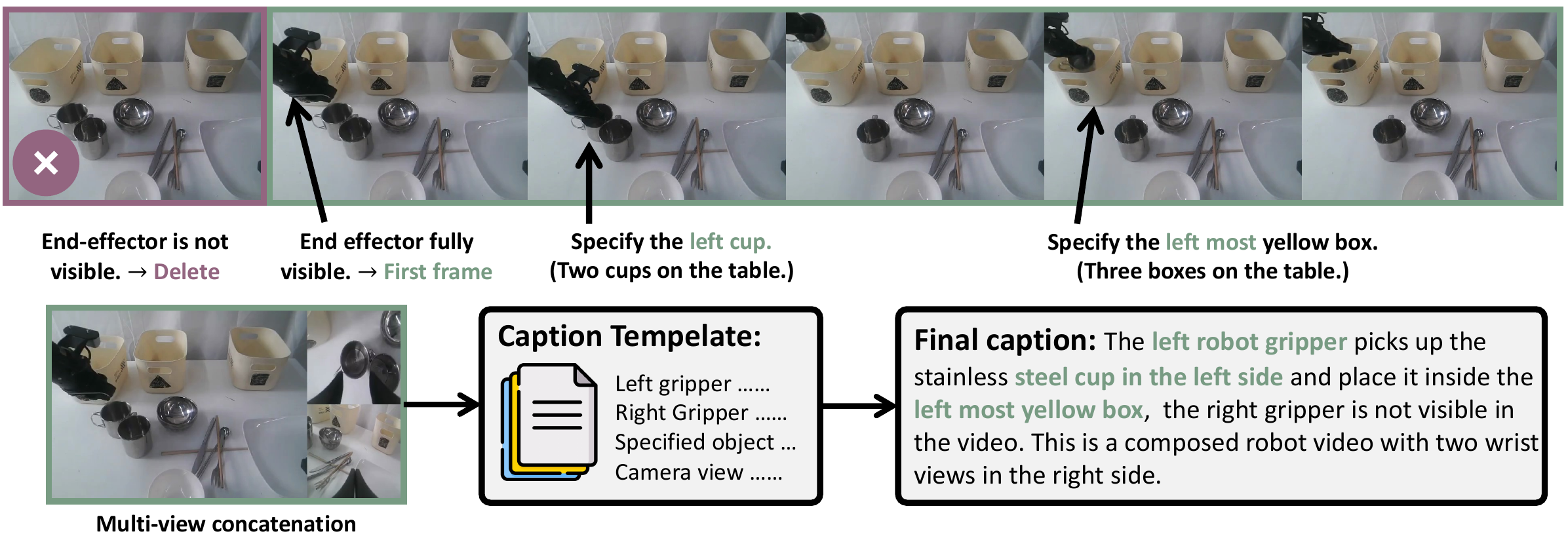}
    \caption{An example of our data processing pipeline. The central objective is to reduce uncertainty in future prediction, thereby encouraging the model to learn a consistent mapping from conditioning information to sub-task trajectories. To this end, each caption includes detailed task descriptions, explicit end-effector identification, a description of the visible target object, and camera-view changes.}
    \label{fig:fig2}
    \vspace{-4mm}
\end{figure}

\textbf{Video Captioning.}
After segmentation, we use a VLM to generate detailed captions for each clip according to a predefined template. Our captioning pipeline aims to reduce uncertainty in future prediction and encourage consistent mappings from semantic descriptions to visual trajectories. Figure~\ref{fig:fig2} illustrates an example. 
Specifically, we address four major sources of uncertainty:
\textit{(1) Detailed Task Description.}
We describe how the end effector completes the task, including the sequence and manner of manipulation.
\textit{(2) Explicit End-Effector Identification.}
The active end effector must be visible in the first frame; otherwise, we trim the clip to begin when it first becomes visible. When multiple end effectors are involved, we describe the motion of each one explicitly.
\textit{(3) Unambiguous Target-Object Specification.}
When multiple identical or similar objects are present, we identify the target using distinctive attributes or spatial relationships. The target object must also be visible in the first frame.
\textit{(4) Camera-View Description.}
We use optical flow to remove egocentric clips with excessive viewpoint changes. For clips with moderate camera motion, a VLM describes the viewpoint changes and the camera views included in the video, and we incorporate this information into the caption.

% \begin{itemize}
% \vspace{-2mm}
%     \item \textit{(1) Detailed Task Description.} We describe how the end effector completes the task, including the sequence and manner of manipulation.
%     \item \textit{(2) Explicit End-Effector Identification.} The active end effector must be visible in the first frame; otherwise, we trim the clip to begin when it first becomes visible. When multiple end effectors are involved, we describe the motion of each one explicitly.
%     \item \textit{(3) Unambiguous Target-Object Specification.}
%     When multiple identical or similar objects are present, we identify the target using distinctive attributes or spatial relationships. The target object must also be visible in the first frame.
%     \item \textit{(4) Camera-View Description.}
%     We use optical flow to remove egocentric clips with excessive viewpoint changes. For clips with moderate camera motion, a VLM describes the viewpoint changes and the camera views included in the video, and we incorporate this information into the caption.
%     \vspace{-2mm}
% \end{itemize}

% After segmentation, we generate detailed captions that describe the task, explicitly identify the active end effectors, unambiguously specify the target objects, and characterize the camera views and motion. Figure~\ref{fig:fig2} illustrates an example, and Appendix~\ref{app_video} provides further details.

\begin{wrapfigure}{t}{0.48\textwidth}
    \vspace{-12pt}
    \centering
    \includegraphics[width=\linewidth]{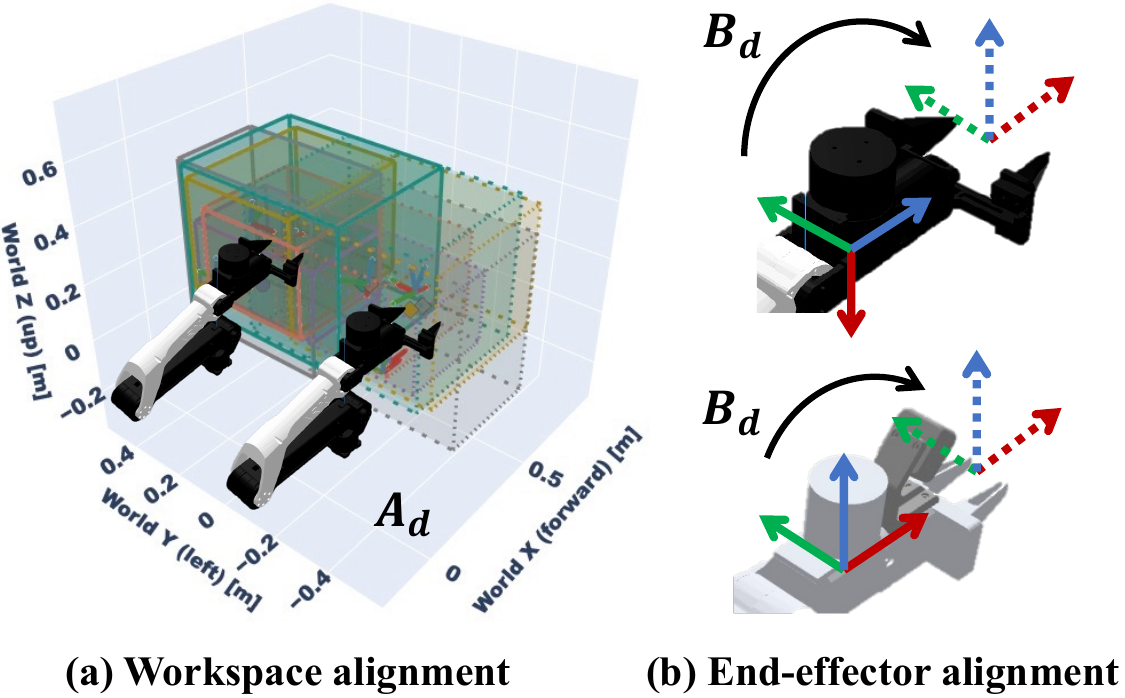}
    \caption{(a) Bounding boxes show the aligned workspaces of different datasets. (b) Aligned end-effector coordinate frames.}
    \label{fig:coordinate_alignment}
    \vspace{-12pt}
\end{wrapfigure}

\textbf{Unified Action Space and Unified End-effector Coordinate Systems.}\label{data:action}
Different robot datasets often adopt inconsistent coordinate systems and camera viewpoints, resulting in conflicting action representations for visually similar motions. For example, a leftward motion may correspond to the positive \(x\)-direction in one dataset but the negative \(y\)-direction in another, making it difficult for the model to learn the semantic meaning of each action dimension.

We focus on egocentric bimanual datasets, including ALOHA-style,
humanoid-style, and human egocentric datasets. These datasets
account for more than \(80\%\) of the total data and share a
similar bimanual structure and camera viewpoint.
We explicitly align their coordinate systems at two levels:
(1) \textit{workspace alignment}, which applies a world-frame
transformation so that different robots have comparable
end-effector workspaces; and (2) \textit{end-effector alignment},
which applies a local transformation to standardize end-effector
origins and axis orientations. 

% Such alignment is necessary because different datasets adopt different end-effector conventions. For example, some place the eef origin between the fingertips, whereas others place it at the gripper mounting link. Similarly, some align the \(x\)-axis with the gripper direction, while others define it perpendicular to that direction.

For an arm dataset \(d\) , let
\(T_d = {}^{W_d}T_{E_d} \in SE(3)\) denote the original
end-effector pose, where \(W_d\) and \(E_d\) are the dataset's
world and end-effector frames, respectively.
We define the workspace alignment as
\(A_d = {}^{\bar{W}}T_{W_d}\) and the end-effector alignment as
\(B_d = {}^{E_d}T_{\bar{E}}\), where \(\bar{W}\) and \(\bar{E}\)
denote the canonical world and end-effector frames.
The aligned end-effector pose is given by $\widetilde{T}_d = A_d T_d B_d,
    \label{eq:coordinate_alignment}$.

\textbf{Unified T-shape Multi-view Input across Datasets.} 
Robotics datasets commonly contain observations from multiple camera views. We arrange them into a T-shaped composite image, with the primary view on the left and two auxiliary views stacked vertically on the right, as illustrated in the bottom-left panel of Figure~\ref{fig:fig2}. When fewer than two auxiliary views are available, we fill the missing slots with black placeholder images.

% \begin{equation}
%     \widetilde{T}_d = A_d T_d B_d,
%     \label{eq:coordinate_alignment}
% \end{equation}
% where \(A_d \in SE(3)\) aligns the world coordinate system
% through left multiplication, and \(B_d \in SE(3)\) aligns
% the end-effector origin and axis orientations through
% right multiplication.
% Such alignment is necessary because different datasets adopt
% different end-effector conventions. For example, some place
% the end-effector origin between the fingertips, whereas others
% place it at the gripper mounting link. Similarly, some align
% the \(x\)-axis with the gripper direction, while others define
% it perpendicular to that direction. The local transformation
% \(B_d\) accounts for both the origin offset and the difference
% in axis orientations.

\section{VPP2: A Generalist Policy with Zero-shot Capability}

\begin{figure}[t]
    \centering
    \includegraphics[width=0.95\linewidth]{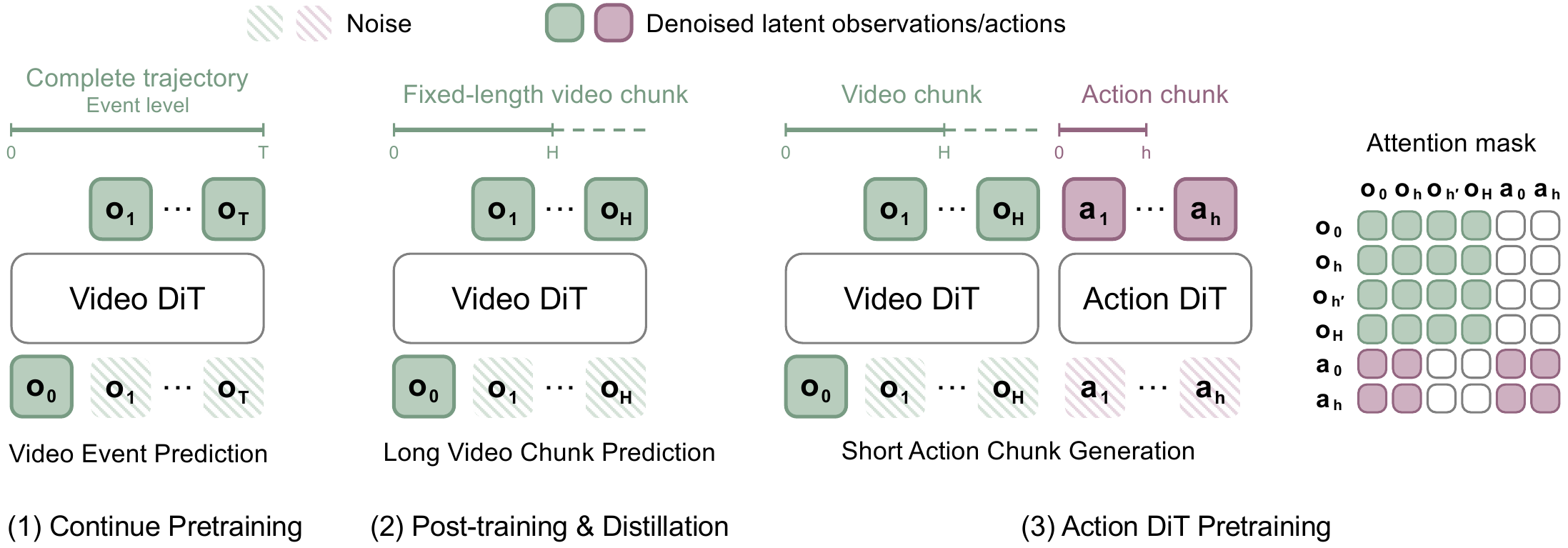}
    \caption{The VPP2 training pipeline is designed to maximize generalization. Stage 1 uses large-scale, event-level video pretraining to learn a generalizable video model for manipulation. Stage 2 post-trains and distills this model to predict long-horizon video chunks spanning 8 seconds in a single forward pass, taking approximately 0.1 seconds. Finally, Stage 3 trains an action expert to generate 2-second action chunks conditioned on the one-step video latents.}
    \label{fig:fig4}
\end{figure}

\textbf{Overview.}
Our large-scale, diverse, and densely annotated dataset enables us to train a policy with strong generalization capabilities. VPP2 builds upon the pretrained Wan2.1-I2V-14B model~\citep{wan2025wan} and introduces an action expert via a mixture-of-transformers (MoT) architecture~\citep{liang2025mixture}. Our training objective is twofold: we first adapt the video model to produce generalizable future predictions at real-time inference speed for open-ended manipulation tasks, and then train the action expert conditioned on the KV cache of the video model. To maximize policy generalization, we organize training into the stages summarized in Figure~\ref{fig:fig4}.

\subsection{Video Prediction Model Training Pipeline}

\textbf{Stage 1: Event-level Video Model Continued Pre-training.}
We first adapt a pretrained video generation model to the manipulation
domain through event-level continued pre-training.
Each training example covers a complete manipulation subtask, aligning
the subtask description with the corresponding visual evolution from
the initial observation to subtask completion.

Given a demonstration $\tau = (o_0, o_1, \ldots, o_T)$,
we uniformly sample $N$ future frames across the entire event:
\begin{equation}
    \mathbf{y}_{\mathrm{event}}
    =
    \left( o_0,
        o_{\lfloor T/N \rfloor},
        o_{\lfloor 2T/N \rfloor},
        \ldots,
        o_T
    \right),
\end{equation}
where $T$ is the final frame index and $N$ excludes the initial
conditioning frame $o_0$.
Let $\mathbf{x}_1 = \mathcal{E}(\mathbf{y}_{\mathrm{event}})$
denote the corresponding video latent representation, where
$\mathcal{E}$ is the video encoder.
We construct the interpolated latent as
\begin{equation}
    \mathbf{x}_s
    =
    (1-s)\boldsymbol{\epsilon} + s\mathbf{x}_1,
    \qquad
    \boldsymbol{\epsilon} \sim \mathcal{N}(\mathbf{0}, \mathbf{I}),
    \quad s \in [0,1],
\end{equation}
where $s$ denotes flow time, with $s=0$ corresponding to noise and
$s=1$ to data.
The model is optimized with the flow matching objective~\citep{lipman2023flow}
\begin{equation}
    \mathcal{L}_{\mathrm{FM}}(\theta)
    =
    \mathbb{E}_{(\mathbf{x}_1,c),\,\boldsymbol{\epsilon},\,s}
    \left[
        \left\|
            v_\theta(\mathbf{x}_s,s,c)
            -
            (\mathbf{x}_1-\boldsymbol{\epsilon})
        \right\|_2^2
    \right],
    \label{eq:video_fm}
\end{equation}
where $v_\theta$ predicts the flow velocity and $c$ includes the
subtask instruction and initial observation $o_0$.

In this stage, we predict 49 frames at a resolution of \(416 \times 240\) and continue pre-training Wan2.1-I2V-14B for $30{,}000$ optimization
steps with batch size $1{,}024$ and learning rate 1e-5.
Our experiments in Sec.~\ref{exp_video}  show that the resulting model achieves strong
instruction following on manipulation tasks, outperforming the 64B Cosmos3-Super-Image2Video model~\citep{agarwal2026cosmos}.
% Our experiments show that the resulting model achieves better
% instruction following on manipulation tasks than the base
% Wan2.1-I2V-14B model~\citep{wan2025wan}.
% Despite having only 14B parameters, VPP2 also outperforms the 64B
% Cosmos3-Super-Image2Video model~\citep{agarwal2026cosmos} on our manipulation
% instruction-following evaluation.

\medskip
\noindent
\textbf{Stage 2: Chunk-level Video Model Post-training.}
Event-level continued pre-training enables the model to generalize well to manipulation tasks.
However, sampling a fixed number of frames from events of different
durations produces variable temporal intervals between predicted
frames, complicating their alignment with downstream actions.
We therefore further post-train the model to predict a fixed-duration
future chunk, establishing a consistent temporal scale for
action learning. In experiments, we predict 2 seconds for human-hand manipulation and 8 seconds for robot manipulation.

Let $H$ denote the prediction horizon in seconds and $f$ the
dataset frame rate in frames per second (fps).
The nominal interval between sampled frames is
$\delta = fH/N$, measured in dataset frames.
For a chunk starting at frame $t$, the target video is
\begin{equation}
    \mathbf{y}_{\mathrm{chunk}}^{(t)}
    =
    \left( o_t,
        o_{t+\lfloor \delta \rfloor},
        o_{t+\lfloor 2\delta \rfloor},
        \ldots,
        o_{t+\lfloor N\delta \rfloor}
    \right),
\end{equation}
with conditioning inputs $c$ containing the observation $o_t$
and the corresponding subtask instruction.
The index $t+\lfloor fH \rfloor$ is capped at the final frame index $T$, and we optimize the same flow matching
objective in Eq.~\eqref{eq:video_fm}, with
$\mathbf{x}_1 = \mathcal{E}(\mathbf{y}_{\mathrm{chunk}}^{(t)})$.

\medskip
\noindent
\textbf{Stage 2: Consistency Distillation.}\label{consistency}
We distill the chunk-level video model for single-step generation~\citep{song2023consistency}
by enforcing consistent terminal predictions at adjacent flow times:
\begin{equation}
    \mathcal{L}_{\mathrm{CD}}
    =
    \mathbb{E}
    \left[
        \lambda(s,s')
        \left\|
            F_\theta(\mathbf{x}_s,s,c)
            -
            \operatorname{stopgrad}\left(
                F_{\bar{\theta}}(\widehat{\mathbf{x}}_{s'},s',c)
            \right)
        \right\|_2^2
    \right],
\end{equation}
where $0 \leq s < s' \leq 1$,
$\widehat{\mathbf{x}}_{s'}$ is obtained by integrating the frozen
teacher flow from $(\mathbf{x}_s,s)$ to $s'$, and
$\bar{\theta}$ denotes the EMA student parameters.
The student satisfies $F_\theta(\mathbf{x},1,c)=\mathbf{x}$.
To emphasize single-step generation, we sample $s'=1$ with
probability $0.5$;
otherwise, we sample $s'$ from the remaining flow times.
At inference, the video latent is generated in a single forward
pass as $F_\theta(\boldsymbol{\epsilon},0,c)$.

\subsection{Action Modeling}
\textbf{Stage 3: Action Pretraining across Datasets.}
After training a generalizable video model, we train the action expert on our processed bimanual datasets with unified workspace and end-effector coordinate systems. The action expert is a 0.9B-parameter diffusion transformer (DiT)~\citep{peebles2023scalable} with a standard MoT architecture. Since the action expert is newly initialized, we freeze the base parameters of the video DiT during the early stages of action training and adapt the video backbone using only LoRA~\citep{hu2022lora}. This strategy helps preserve the pretrained video representations while limiting disruption from action-training gradients.

\textbf{Inference Latency.}
We generate video efficiently by reducing the video input to \(17\times416 \times 240\) during post-training and distilling the video model for one-step generation, as described in Sec.~\ref{consistency}. With bfloat16 precision and torch.compile, the video model takes approximately 0.12 seconds. The action expert then generates an action chunk conditioned on the video latents using five denoising steps, taking approximately 0.1 seconds. The total latency per action chunk is therefore approximately 0.22 seconds.

% The original model processes videos at a spatiotemporal resolution of \(81 \times 832 \times 480\) and requires several seconds even for a single forward pass, making it impractical for real-time robotic systems. Rather than pursuing pixel-level fidelity, we prioritize reliable instruction following and informative future predictions that can effectively guide the action expert. Accordingly, we reduce the sequence length to 17 frames, comprising one observed frame and 16 predicted frames, and lower the spatial resolution to \(416 \times 240\). With this configuration, a forward pass takes approximately 0.12 seconds on a single GPU, satisfying the latency requirements of real-time policy inference.

\subsection{VLM for High-level Planning.}
\textbf{Sub-task Planning.} To tackle long-horizon tasks and handle ambiguous human instructions, we adopt a hierarchical architecture in which a VLM generates subtask plans~\citep{shi2025hi}. The VLM handles high-level reasoning, semantic understanding, and memory, allowing VPP2 to focus on mapping explicit instructions to trajectories. In practice, the VLM can be deployed locally or accessed through an API to a frontier model.

\textbf{Prompt Enhancement.}
VLM planning also generates detailed subtask descriptions that match the caption template used during training. Prior work on video generation likewise highlights the importance of detailed captions for generation quality~\citep{wan2025wan,yang2025cogvideox}.

% Requires: \usepackage{booktabs}
% \begin{table}[h]
%     \centering
%     \small
%     \setlength{\tabcolsep}{4pt}
%     \renewcommand{\arraystretch}{1.15}
%     \begin{tabularx}{\linewidth}{@{}llXr@{}}
%         \toprule
%         Data Type & Embodiment Type & Data Sources & Total Time \\
%         \midrule
%         \multirow{3}{*}{Robot}
%         & Single-arm
%         & OXE, RoboMIND, DROID, RH20T, Molmoact
%         & 3,624 h \\
%         & Dual-arm
%         & AgibotWorld-Beta, RoboCOIN, RDT, GM100, ABC-130k, Self-collected Aloha
%         & 10,307 h \\
%         & Mobile \& humanoid
%         & InternData-A1, Galaxea Open-World
%         & 868 h \\
%         \midrule
%         \multirow{2}{*}{Human}
%         & \multirow{2}{*}{Human hands}
%         & Open-source: EgoDex, Ego4D, Epic-Kitchen
%         & 4,599 h \\
%         & & Self-collected & $\approx10,000$h \\
%         \midrule
%         General Video
%         & Human hands
%         & Open-vid, Panda-70M
%         & --- \\
%         \bottomrule
%     \end{tabularx}
%     \caption{Placeholder Caption}
%     \label{table:video_source}
% \end{table}

\begin{figure}[t]
    \centering
    \includegraphics[width=1.0\linewidth]{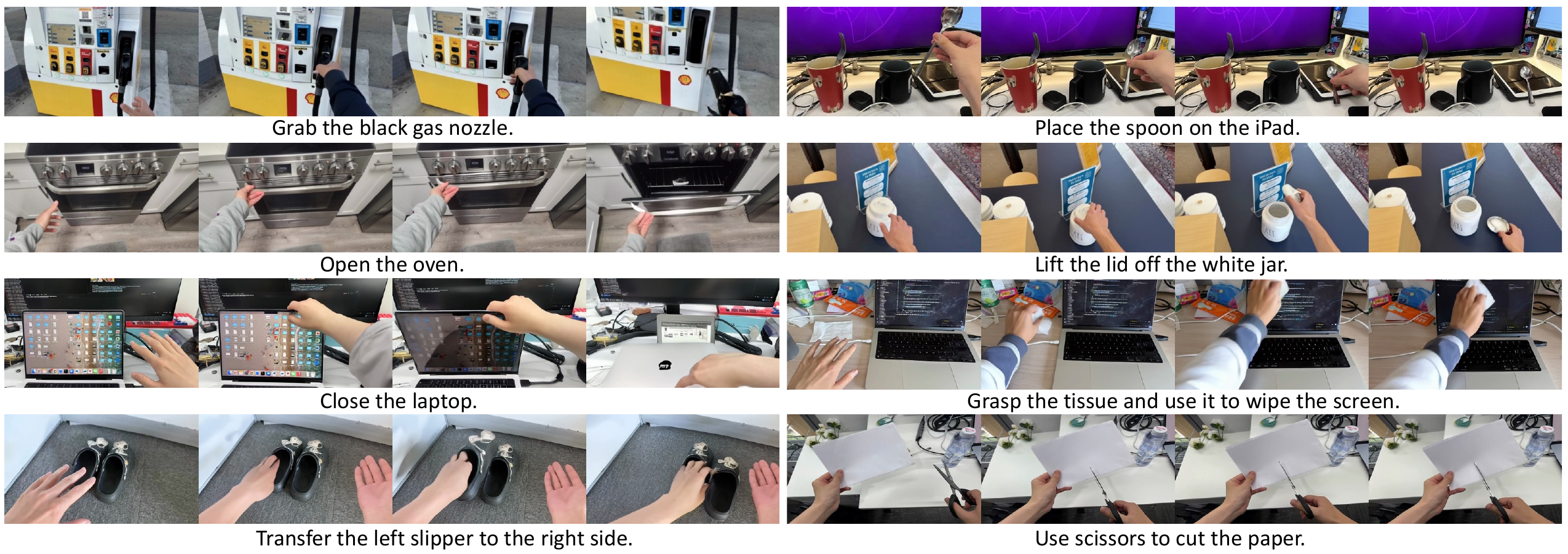}
    \includegraphics[width=1.0\linewidth]{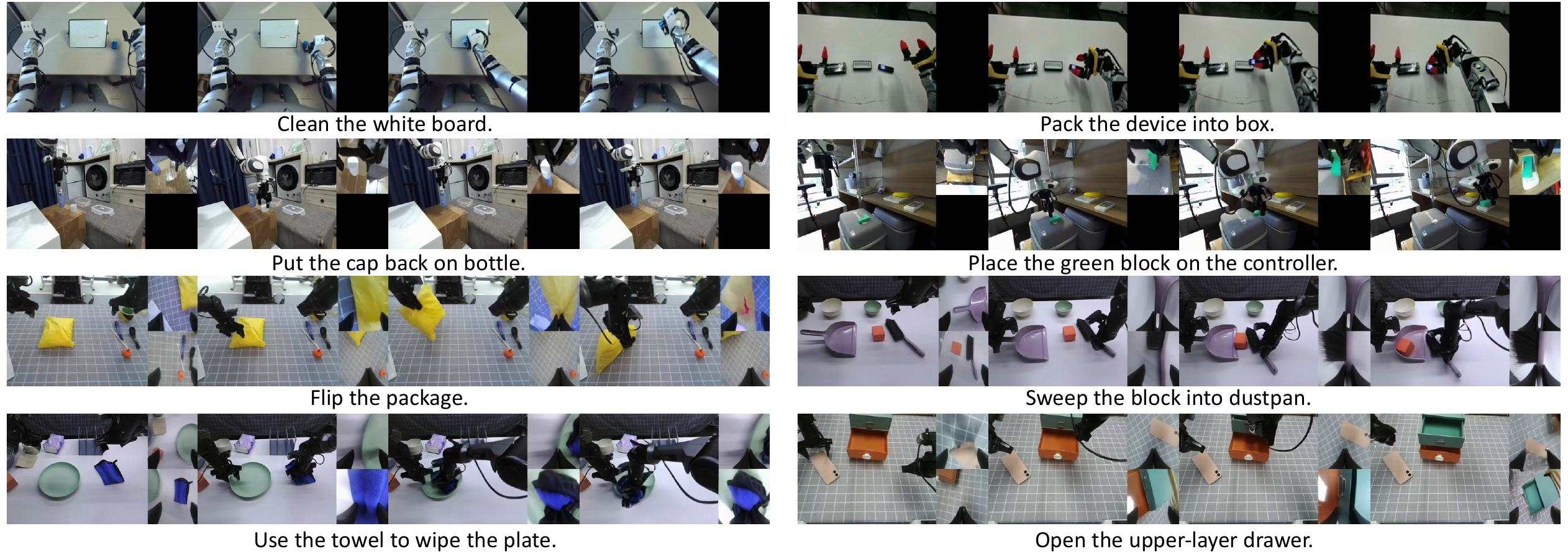}
    \caption{VPP2 video predictions for human-hand and robot manipulation. Trained on large-scale, diverse manipulation datasets, VPP2 generalizes across human hands and a wide range of robot embodiments. For robot manipulation, VPP2 jointly predicts one to three camera views arranged in a T-shaped composite. Due to space constraints, single-step video predictions after distillation are shown in Figure~\ref{fig:video_app} in the appendix.}
    \vspace{-4mm}
    \label{fig:fig6}
\end{figure}

% \begin{figure}[t]
%     \centering
%     \includegraphics[width=1.0\linewidth]{figures/fig7.pdf}
%     \caption{VPP2 video predictions on robot manipulation video prediction. By training on large diverse robot datasets, VPP2 support different robot embodiment and different number of camera views (main-view is put in the left and two possible auxiliary views is put on the right side).}
%     \label{fig:fig7}
% \end{figure}

\section{Experiments}
In this section, we conduct experiments to answer the following questions:
(1) How does the motion quality of VPP2's video predictions compare with that of state-of-the-art video foundation models on manipulation tasks?
(2) How broad are VPP2's zero-shot manipulation capabilities?
(3) How well does VPP2 perform after domain-specific post-training?

\subsection{Video Prediction Quality Analysis}\label{exp_video}

\textbf{Video Prediction for Human and Robot Manipulation.}
Figure~\ref{fig:fig6} visualizes predictions for human-hand manipulation and diverse robot embodiments from the pretrained VPP2 model. To assess generalization, we randomly sample test cases from prior work~\citep{chen2025large,zhang2026qwen} and take photographs in office environments and robot workspaces, pairing them with freely chosen manipulation instructions.
Trained on data spanning more than 20 robot embodiments, VPP2 also generalizes to unseen embodiments and tasks, as shown in Figure~\ref{fig:fig6}. VPP2 also supports joint prediction across one to three camera views, with the primary view placed on the left and up to two auxiliary views stacked on the right.

\textbf{Quantitative Comparisons.}
We compare VPP2 with four video foundation models and one ablation variant: (1) Wan-2.1-I2V-14B-480p~\citep{wan2025wan}, our base model; (2) LVP~\citep{chen2025large}, which also adapts Wan-2.1-I2V-14B-480p to manipulation tasks; (3) Cosmos3-Nano-16B, which is extensively trained on robot manipulation data; (4) Cosmos3-Super-Image2Video-64B, the strongest model in the Cosmos 3 family~\citep{agarwal2026cosmos}; and (5) VPP2-Fixed-Step, an ablation of our method that predicts a fixed-duration future chunk rather than a complete event, potentially misaligning the predicted future with the instruction describing the full event.

\begin{figure}[t]
\centering

% Left: table
\begin{minipage}[t]{0.32\linewidth}
\vspace{0pt}
\centering
\small
\setlength{\tabcolsep}{3pt}
\renewcommand{\arraystretch}{0.9}

\begin{tabularx}{\linewidth}{@{}l
    >{\centering\arraybackslash}X
    >{\centering\arraybackslash}X@{}}
\toprule
Method & Human & Robot \\
\midrule
Wan2.1-14B  & 0.32 & 0.04 \\
LVP        & 0.62 & 0.46 \\
Cosmos3-16B & 0.44 & 0.32 \\
Cosmos3-64B & 0.70 & 0.78 \\
VPP2-Fixed-Step      & 0.48 & 0.42 \\
VPP2 (ours)      & \textbf{0.80} & \textbf{0.90} \\
\bottomrule
\end{tabularx}

\captionof{table}{Instruction following success rates.}
\label{table:video_quality}
\end{minipage}
\hfill
% Right: figure
\begin{minipage}[t]{0.66\linewidth}
\vspace{0pt}
\centering
\includegraphics[width=\linewidth]{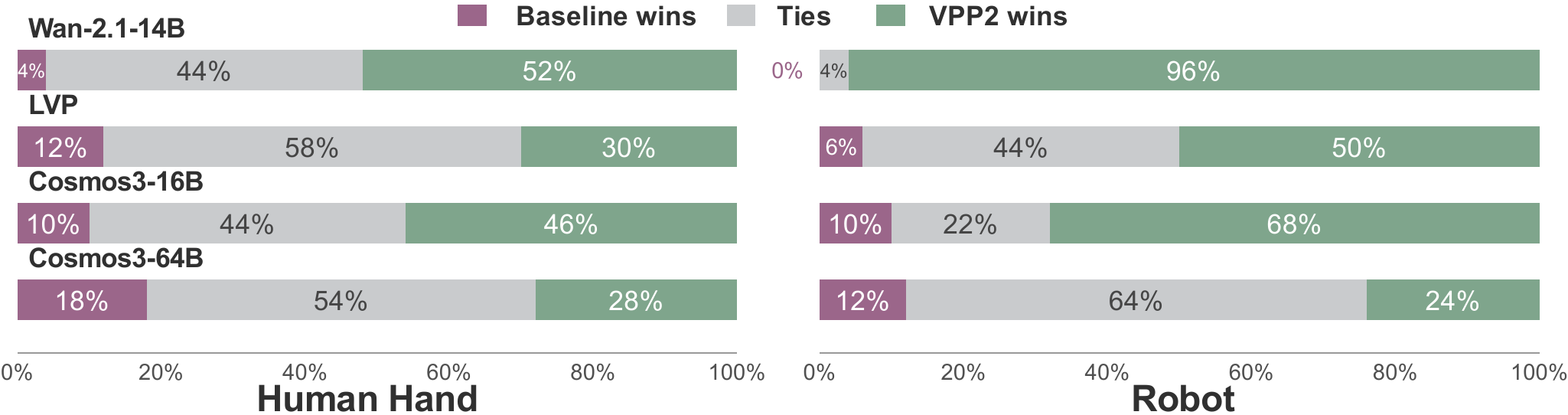}

\caption{Win rates of the pretrained VPP2 model against different baselines on video prediction tasks.}
\label{fig:fig8}
\end{minipage}

\end{figure}

\begin{figure}[t]
    \centering
    \vspace{-2mm}
    \includegraphics[width=1.0\linewidth]{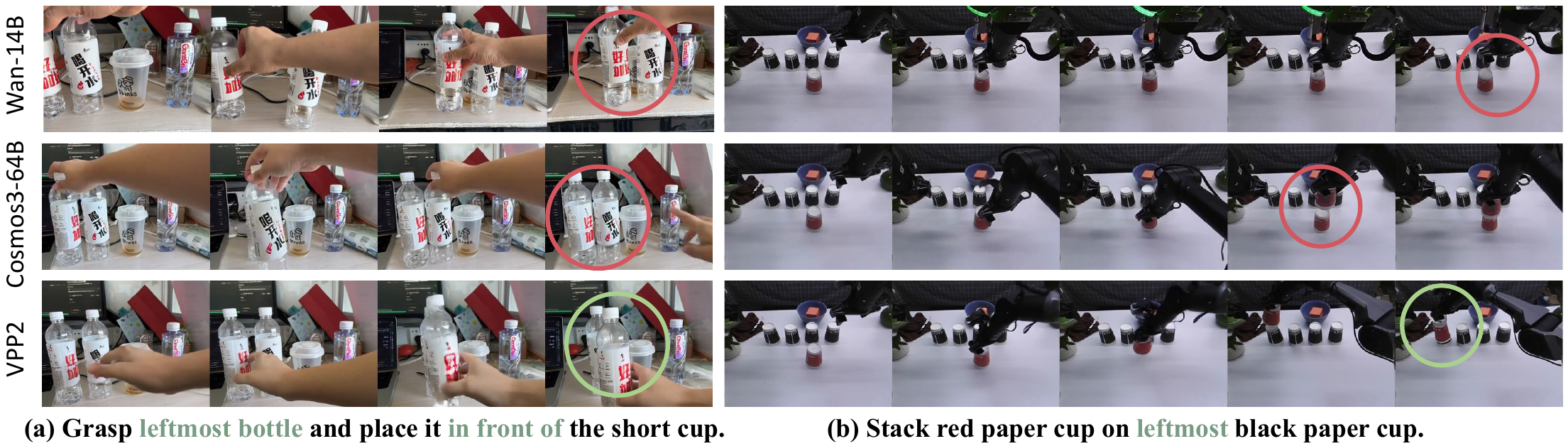}
    \caption{Comparison of instruction-following capabilities among VPP2, Wan-14B, and Cosmos3-64B. VPP2 demonstrates better instruction following on complex tasks requiring spatial understanding.}
    \label{fig:fig5}
    \vspace{-4mm}
\end{figure}

We randomly collect 50 image--instruction pairs each for human-hand and robot manipulation and generate video predictions for every pair using each model. We first use a GPT-based evaluator to assess instruction-following success rates, reported in Table~\ref{table:video_quality}. Human evaluators then compare pairs of predictions to determine which better follows the instruction, yielding VPP2's pairwise win rates against each baseline in Figure~\ref{fig:fig8}. VPP2's advantages are most pronounced in complex manipulation scenarios requiring spatial understanding, as illustrated in Figure~\ref{fig:fig5}. We hypothesize that training with detailed captions encourages more accurate mappings from instructions to trajectories, contributing to these improvements.

% Add to the preamble
% \usepackage{booktabs,tabularx,caption,graphicx}

% Main content

\subsection{Policy Performance Analysis}
\begin{figure}[t]
    \centering
    \includegraphics[width=1.0\linewidth]{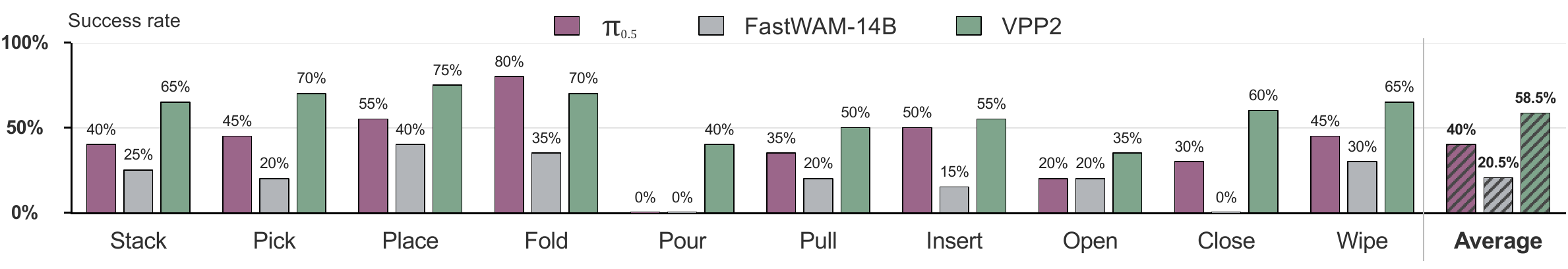}
    \caption{Zero-shot success rates on the real-world ALOHA across 10 randomly selected task categories. For a fair comparison, we finetune $\pi_{0.5}$ and FastWAM on all ALOHA datasets exposed in VPP2’s training data and use the 14B variant of FastWAM.}
    \label{fig:fig9}
    \vspace{-2mm}
\end{figure}

\textbf{Zero-shot Performance on Real-world Aloha Robot.} After pretraining a strong video foundation model for manipulation, we post-train and distill it into a fast, single-step generator to support action learning.
With its generalizable video predictions, VPP2 demonstrates strong zero-shot capabilities across open-ended manipulation tasks. We deploy VPP2 directly on a robot embodiment seen during training, without task-specific fine-tuning. For comparison, we train $\pi_{0.5}$~\citep{intelligence2025pi_} and the Wan-14B version of Fast-WAM~\citep{yuan2026fastwam} on the same data used in VPP2 training.
We evaluate models on 10 different categories of zero-shot tasks and present results in Figure~\ref{fig:fig9}. VPP2 achieves an average success rate of 58.5\%, compared with 40.0\% for $\pi_{0.5}$ and 20.5\% for Fast-WAM, and performs best in 9 of the 10 categories.

\begin{table*}[t]
\centering

\setlength{\tabcolsep}{5.0pt}
\renewcommand{\arraystretch}{1.0}

\resizebox{0.9\textwidth}{!}{
\begin{tabular}{l c ccc cccc}
\toprule
& \multicolumn{1}{c}{\textbf{LIBERO-ID}}
& \multicolumn{3}{c}{\textbf{LIBERO-Pro}}
& \multicolumn{4}{c}{\textbf{LIBERO-OOD}} \\
\cmidrule(lr){2-2}
\cmidrule(lr){3-5}
\cmidrule(lr){6-9}

\textbf{Method}
& \textbf{Overall}
& \textbf{Position}
& \textbf{Task}
& \textbf{Overall}
& \textbf{Spatial}
& \textbf{Object}
& \textbf{Goal}
& \textbf{Overall} \\
\midrule

$\pi_0$~\citep{black2024pi_0}
& 94.2
& 0.5 & 0.0 & 0.3
& 0.7 & 0.3 & 4.3 & 1.7 \\

$\pi_{0.5}$~\citep{intelligence2025pi_}
& 96.9
& 20.8 & 1.3 & 11.0
& 36.7 & 2.3 & 41.7 & 26.8 \\

MolmoAct~\citep{lee2025molmoact}
& 86.6
& 1.5 & 1.5 & 1.5
& -- & -- & -- & -- \\

X-VLA~\citep{zheng2026xvla}
& 98.1
& 0.8 & 6.8 & 3.8
& -- & -- & -- & -- \\

AtomVLA~\citep{sun2026atomvla}
& 97.0
& 7.3 & 5.3 & 6.3
& -- & -- & -- & -- \\

\midrule

Cosmos-Policy~\citep{kim2026cosmospolicy}
& 98.5
& -- & -- & --
& 30.7 & 0.3 & 0.7 & 10.5 \\

Fast-WAM~\citep{yuan2026fastwam}
& 97.6
& -- & -- & --
& 12.7 & 0.0 & 15.7 & 9.4 \\

DiT4DiT~\citep{ma2026dit4dit}
& 98.6
& -- & -- & --
& 9.0 & 0.0 & 10.3 & 6.4 \\

Temporal Ratio~\citep{mishra2026understanding}
& 94.0
& -- & -- & --
& 58.6 & 40.0 & 80.3 & 59.4 \\

\midrule

\textbf{VPP2 (Ours)}
& \textbf{98.8}
& \textbf{42.3}
& \textbf{47.8}
& \textbf{45.0}
& \textbf{59.3}
& \textbf{44.5}
& \textbf{87.7}
& \textbf{63.9} \\

\bottomrule
\end{tabular}
}
\caption{
Post-training on LIBERO-ID, LIBERO-Pro, and LIBERO-OOD benchmarks.
All models are trained exclusively on the four standard LIBERO suites~\citep{liu2023libero} and are evaluated under three complementary settings.
% For LIBERO-Pro, Position and Task are macro-averaged over the corresponding position and task perturbations across the evaluated suites.
% LIBERO-OOD evaluates compositional generalization along the spatial, object, and goal dimensions.
% Overall denotes the average success rate within each benchmark.
}
\label{tab:libero_generalization}
\vspace{-2mm}
\end{table*}

\begin{table}[t]
\centering
\setlength{\tabcolsep}{4pt}
\resizebox{\textwidth}{!}{
\begin{tabular}{lcccccccc}
\toprule
 & $\pi_0$ & $\pi_{0.5}$ & X-VLA & Fast-WAM & OpenWAM-$\alpha$ & Xiaomi-Robotics-1 & GPT-6-Astra & \textbf{VPP2 (Ours)} \\
\midrule
Avg.\ Score & 3.48 & 11.44 & 10.13 & 3.48 & 17.18 & 20.07 & 28.97 & \textbf{39.26} \\
Success Rate (\%) & 1.53 & 6.93 & 6.52 & 2.03 & 11.92 & 13.93 & 22.48 & \textbf{32.26} \\
\bottomrule
\end{tabular}
}
\caption{Post-training results on the RoboDojo simulation benchmark. Average score captures partial task progress, and success rate measures binary task completion; both are averaged over the five capability dimensions. Baseline results are taken from the official leaderboard.}
\label{tab:robodojo}
\end{table}

\textbf{LIBERO-ID, LIBERO-Pro, and LIBERO-OOD Benchmarks.}
All models are trained exclusively on the four standard LIBERO suites~\citep{liu2023libero} and are evaluated under three complementary settings. LIBERO-ID measures standard in-distribution manipulation performance. To assess generalization beyond the training distribution, we further evaluate on LIBERO-Pro~\citep{zhou2025liberopro} and LIBERO-OOD~\citep{li2025vlas} without additional training. Following HarnessVLA~\citep{zhang2026harness}, we evaluate the Position and Task perturbations of LIBERO-Pro, which test generalization to changes in object positions and task specifications, respectively. We further follow~\citep{mishra2026understanding} to evaluate compositional generalization on LIBERO-OOD, which recombines familiar objects, layouts, and goals into unseen task configurations along spatial, object, and goal dimensions. Per-suite LIBERO-ID results are provided in Appendix~\ref{app_libero}.

As shown in Table~\ref{tab:libero_generalization}, while most baselines exceed 94\% on LIBERO-ID, this in-distribution performance does not transfer to generalization. On LIBERO-Pro, VLA baselines drop to at most 11.0\% overall, whereas VPP2 reaches 45.0\%, with the largest gain on the Task perturbation (47.8\% vs.\ 6.8\%), indicating that VPP2 follows the given instruction rather than replaying memorized trajectories. On LIBERO-OOD, video-based policies such as Cosmos-Policy, Fast-WAM, and DiT4DiT reach at most 10.5\%, while VPP2 achieves 63.9\%, also surpassing Temporal Ratio~\citep{mishra2026understanding}, which specifically targets this generalization gap. We attribute these gains to next-event video prediction with detailed captions, which establishes consistent mappings from instructions to future trajectories.

\textbf{RoboDojo Benchmark.}
RoboDojo~\citep{chen2026robodojo} is a unified sim-and-real benchmark for evaluating generalist robot manipulation policies. Its simulation benchmark contains 42 bimanual tasks covering five capability dimensions: generalization, memory, precision, long-horizon execution, and open-vocabulary instruction following. We follow the official training and evaluation protocol and report both the success rate and the average score, which captures partial task progress.

To support memory-dependent tasks, we augment the model input with the initial frame of the episode ($t=0$) and eight past frames sampled at one-second intervals. Since this modification changes the model's input structure, we initialize RoboDojo-specific post-training from the Stage~2 checkpoint before consistency distillation. We first train the video model alone for $10{,}000$ steps, then jointly train the video model and action expert. Throughout training, the video noise level is fixed at its maximum, and video prediction requires only a single forward pass at inference. During joint training, the action flow timestep is randomly sampled from $[0,1]$.

As shown in Table~\ref{tab:robodojo}, VPP2 achieves an average score of 39.26 and a success rate of 32.26\%, achieving state-of-the-art performance.\footnote{Baseline results are taken from the official RoboDojo simulation leaderboard as of September 2026.} It outperforms representative VLAs such as $\pi_{0.5}$~\citep{intelligence2025pi_}, X-VLA~\citep{zheng2026xvla}, and Xiaomi-Robotics-1~\citep{guo2026xiaomirobotics1}, world action models such as Fast-WAM~\citep{yuan2026fastwam} and OpenWAM-$\alpha$~\citep{wang2026openwam}, and the frontier foundation model GPT-6-Astra~\citep{zhang2026astra} (28.97 / 22.48\%). Per-dimension results are provided in Appendix~\ref{app_robodojo}.

\textbf{VPP2 with Subtask Planning.}
The VPP2 results in Table~\ref{tab:robodojo} use full-task instructions without subtask planning. We further investigate VLM-based subtask planning on five selected RoboDojo task groups covering object classification, block stacking, block swapping, mahjong, and tic-tac-toe. To train the subtask-conditioned policy, we segment demonstrations and pair each segment with a subtask instruction. At test time, a vision-language model (VLM) uses visual observations and execution feedback to select the next subtask for VPP2 to execute.

\begin{figure}[t]
    \centering
    \vspace{-3mm}
    \includegraphics[width=\linewidth]{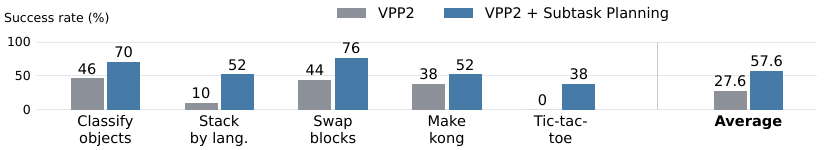}
    \caption{Success rates with and without VLM-based subtask planning (50 trials per task). Average is the five-task mean; the baseline checkpoint matches Table~\ref{tab:robodojo}.}
    \label{fig:subtask_planning}
    \vspace{-3mm}
\end{figure}

Figure~\ref{fig:subtask_planning} summarizes the results across the five selected task groups. The evaluated planning configurations achieve an average success rate of 57.6\%, compared with 27.6\% for the baseline. The largest gains occur in language-conditioned block stacking (10\% to 52\%) and tic-tac-toe (0\% to 38\%).

\section{Related Works}
\textbf{Video Foundation Model.}
Video generation has advanced rapidly through latent diffusion models~\citep{blattmann2023align,blattmann2023stable} and large-scale diffusion transformer models~\citep{yang2025cogvideox,wan2025wan,agarwal2025cosmos,agarwal2026cosmos}. Large-scale bidirectional video models provide strong general-purpose video priors; even recent autoregressive generators build on pretrained bidirectional models, using them for initialization or as teachers before causal adaptation and distillation~\citep{yin2025causvid,huang2025selfforcing}. This motivates our strategy of first learning a generalizable video model and then adapting it for efficient prediction. 
% However, general-purpose video pretraining alone does not ensure faithful execution of manipulation instructions or accurate prediction of task-relevant object interactions from the current observation~\citep{chen2025large,zhang2026qwen}. VPP2 therefore continues pretraining the bidirectional base model on manipulation videos, predicting complete events to learn consistent instruction-to-trajectory mappings.
Besides the training strategy, prompt enhancement is an important component of video foundation models~\citep{yang2025cogvideox,wan2025wan}. Similarly, we use a VLM to produce detailed, manipulation-specific descriptions that reduce ambiguity in these mappings. After establishing this manipulation-focused video prior, we post-train the model for fixed-horizon prediction and distill it for single-step generation to support efficient action learning.
% For robotic manipulation, however, visual realism alone does not ensure faithful instruction following or predictions grounded in the current observation~\citep{chen2025large,zhang2026qwen}. 

% which rely on detailed captions and prompt enhancement for controllability but are optimized for general-purpose video rather than manipulation. 
% In robotics, video models have been trained as action-conditioned simulators and policy evaluators~(\citealp{yang2023learning,zhu2024irasim,wu2024ivideogpt,jiang2025enerverse,he2025pre,li2025worldeval}; \mbox{\citealp{quevedo2025evaluating}}; \citealp{1xworld}), or as language-conditioned planners and data generators~\citep{chen2025large,jang2025dreamgen,chi2025wow,zhang2026qwen}, which typically require slow multi-step sampling. In contrast, VPP2 adapts a video foundation model into a real-time policy backbone with improved manipulation instruction following.

\textbf{World Action Model.}
World action models (WAMs) leverage video prediction to facilitate robot policy learning. Early approaches first generate future frames and then infer actions through inverse dynamics~\citep{du2024learning,black2023zero,bharadhwaj2024gen2act,liang2024dreamitate,feng2025vidar}, but iterative video generation can incur substantial latency for closed-loop control. Hierarchical approaches also combine video-based motion planning with reactive VLA control~\citep{zhang2026veoact}. More recent approaches condition actions on features from video models~\citep{hu2024video,liao2025genie,jia2025video2act,yan2026svam} or jointly learn visual prediction and action generation~\citep{zhang2025up,guo2024prediction,li2025unified,zhu2025unified,ma2026dit4dit,kim2026cosmospolicy,yan2026robustwam,ye2026dreamzero,li2026lingbotva,bi2026motus,motubrainteam2026motubrain}. Recent work also explores event-grounded world-action learning~\citep{li2026wallwm} and large-scale manipulation-specific pretraining~\citep{agibot2026geact2}. Nevertheless, action learning can compromise pretrained capabilities, including video-model generalization~\citep{mishra2026understanding} and VLM semantic understanding~\citep{zhang2026uam}. VPP2 emphasizes instruction-aligned future prediction, combining event-level pretraining with detailed captions, fixed-horizon post-training, and single-step distillation to provide an efficient video backbone for action learning.

\textbf{Generalizable Robot Policy.}
Vision-language-action (VLA) models transfer pretrained visual and semantic knowledge to robot control through training on diverse robot demonstrations~\citep{brohan2023rt,kim2024openvla,black2024pi_0,intelligence2025pi_,liu2024rdt,wen2025dexvla,zheng2026xvla,lee2025molmoact,zhong2025flowvla}. However, stronger general-purpose vision-language capabilities do not necessarily translate into better downstream control, motivating representations and supervision tailored to embodied tasks~\citep{zhang2026vlm4vla}. Recent work improves spatial grounding through target-region reconstruction and alignment with pretrained 3D representations~\citep{song2025reconvla,li2025spatialforcing}, combines multimodal understanding with future prediction~\citep{zhang2025up}, and grounds latent actions in visual and proprioceptive dynamics~\citep{chen2026villax}. Temporal reasoning is further supported by modeling past and future motion~\citep{lin2025hifvla} and interleaving language reasoning, visual forecasting, and action generation for long-horizon manipulation~\citep{hu2026bagelvla}.

Complementary advances improve action generation and execution through hybrid autoregressive-diffusion modeling~\citep{liu2025hybridvla}, lightweight action adapters~\citep{wang2025vlaadapter}, and hierarchical architectures that couple slowly updated semantic representations with fast control~\citep{zhang2024hirt,chen2025fastinslow}. High-level planners further decompose open-ended instructions into executable subtasks~\citep{shi2025hi,cui2025openhelix}, while online reinforcement learning enables pretrained policies to improve through environment interaction~\citep{guo2025improving}. Despite these advances, robustness to task perturbations and novel instructions remains challenging~\citep{zhou2025liberopro,li2025vlas}. Our approach pursues generalization through manipulation-specific video pretraining, using instruction-aligned future predictions to guide efficient action learning.

\vspace{-1mm}
\section{Conclusion}
\vspace{-1mm}
We presented Video Prediction Policy 2 (VPP2), a world action model that achieves strong zero-shot generalization in both video prediction and action generation. We first continue pretraining the base video model to produce generalizable, instruction-following predictions, then learn a policy while preserving these predictive capabilities. Experiments show that VPP2 follows manipulation instructions more faithfully than substantially larger video foundation models, exhibits strong zero-shot manipulation capabilities on a real-world ALOHA platform, and achieves the highest success rates after post-training on challenging benchmarks. We hope these findings encourage future WAM research to look beyond architectural choices for action modeling and place greater emphasis on learning generalizable video predictions and preserving this generalization during action learning.

% These results suggest that better future prediction leads to better action, and we hope VPP2 encourages future WAMs to treat prediction quality as a first-class objective.

% We identify two factors that limit the generalization of existing WAMs: base video models are not optimized for manipulation, and naively incorporating action learning can degrade their predictive capabilities. To address these issues, VPP2 performs event-level continued pretraining with detailed captions on large-scale, diverse manipulation videos, establishing consistent mappings from instructions to future trajectories. It then post-trains and distills the video model into a single-step visual planner with a fixed prediction horizon, and learns actions through a mixture-of-transformers action expert. 

%\newpage

\section*{Acknowledgments}
We thank Yucheng Hu and Jianke Zhang for insightful discussions that greatly benefited the development of VPP2.

\section*{AI Use Statement}
We used generative AI tools to assist with language polishing, improve the clarity and readability of the manuscript, and draft portions of the paper. In the research process, these tools were used to suggest experimental parameter settings and assist with searches for relevant literature and technical information. We also used AI tools to annotate human and robotic manipulation datasets at scale for model training. The authors evaluated the suggestions and made the final decisions regarding experimental design and parameter selection. All AI-assisted text was reviewed and revised by the authors, and retrieved information was checked against original sources before use. The authors take full responsibility for the final content of the paper, including its data, claims, results, and references.

\section*{Reproducibility Statement}
We describe the model architecture, training objectives, and evaluation protocols in the main text, with additional data processing details provided in the appendix. Detailed training configurations, code, and model checkpoints are available on our \href{https://robert-gyj.github.io/video-prediction-policy-2}{project website}.

\bibliography{iclr2027_conference}
\bibliographystyle{iclr2027_conference}

\newpage
\appendix

\section{Dataset Process Details}\label{app_data}

\subsection{Video Captioning.}\label{app_video}
After segmentation, we use a VLM to generate detailed captions for each clip according to a predefined template. Our captioning pipeline aims to reduce uncertainty in future prediction and encourage consistent mappings from semantic descriptions to visual trajectories. Specifically, we address four major sources of uncertainty:

\textit{(1) Detailed Task Description.}
We describe how the end effector completes the task, including the sequence and manner of manipulation.

\textit{(2) Explicit End-Effector Identification.}
The active end effector must be visible in the first frame; otherwise, we trim the clip to begin when it first becomes visible. When multiple end effectors are involved, we describe the motion of each one explicitly.

\textit{(3) Unambiguous Target-Object Specification.}
When multiple identical or similar objects are present, we identify the target using distinctive attributes or spatial relationships. The target object must also be visible in the first frame.

\textit{(4) Camera-View Description.}
We use optical flow to remove egocentric clips with excessive viewpoint changes. For clips with moderate camera motion, a VLM describes the viewpoint changes and the camera views included in the video, and we incorporate this information into the caption.

\subsection{Unified Action Space}

We focus on egocentric bimanual datasets, including ALOHA-style,
humanoid-style, and human egocentric datasets. These datasets
account for more than \(80\%\) of the total data and share a
similar bimanual structure and camera viewpoint.
We explicitly align their coordinate systems at two levels:
(1) \textit{workspace alignment}, which applies a world-frame
transformation so that different robots have comparable
end-effector workspaces; and (2) \textit{end-effector alignment},
which applies a local transformation to standardize end-effector
origins and axis orientations. 

% Such alignment is necessary because different datasets adopt different end-effector conventions. For example, some place the eef origin between the fingertips, whereas others place it at the gripper mounting link. Similarly, some align the \(x\)-axis with the gripper direction, while others define it perpendicular to that direction.

For an arm dataset \(d\) , let
\(T_d = {}^{W_d}T_{E_d} \in SE(3)\) denote the original
end-effector pose, where \(W_d\) and \(E_d\) are the dataset's
world and end-effector frames, respectively.
We define the workspace alignment as
\(A_d = {}^{\bar{W}}T_{W_d}\) and the end-effector alignment as
\(B_d = {}^{E_d}T_{\bar{E}}\), where \(\bar{W}\) and \(\bar{E}\)
denote the canonical world and end-effector frames.
The aligned end-effector pose is given by
\begin{equation}
    \widetilde{T}_d = A_d T_d B_d,
    \label{eq:coordinate_alignment}
\end{equation}
% where \(A_d \in SE(3)\) aligns the world coordinate system
% through left multiplication, and \(B_d \in SE(3)\) aligns
% the end-effector origin and axis orientations through
% right multiplication.
% Such alignment is necessary because different datasets adopt
% different end-effector conventions. For example, some place
% the end-effector origin between the fingertips, whereas others
% place it at the gripper mounting link. Similarly, some align
% the \(x\)-axis with the gripper direction, while others define
% it perpendicular to that direction. The local transformation
% \(B_d\) accounts for both the origin offset and the difference
% in axis orientations.
\newpage
\section{More Video Prediction Results}
% Additional video prediction results on open-ended tasks. We compare single-step video predictions before and after distillation. Distillation enables high-quality video prediction with just one sampling step, as shown in Figure~\ref{fig:video_app}.

\begin{figure}[h]
    \centering
    \includegraphics[width=0.96\linewidth]{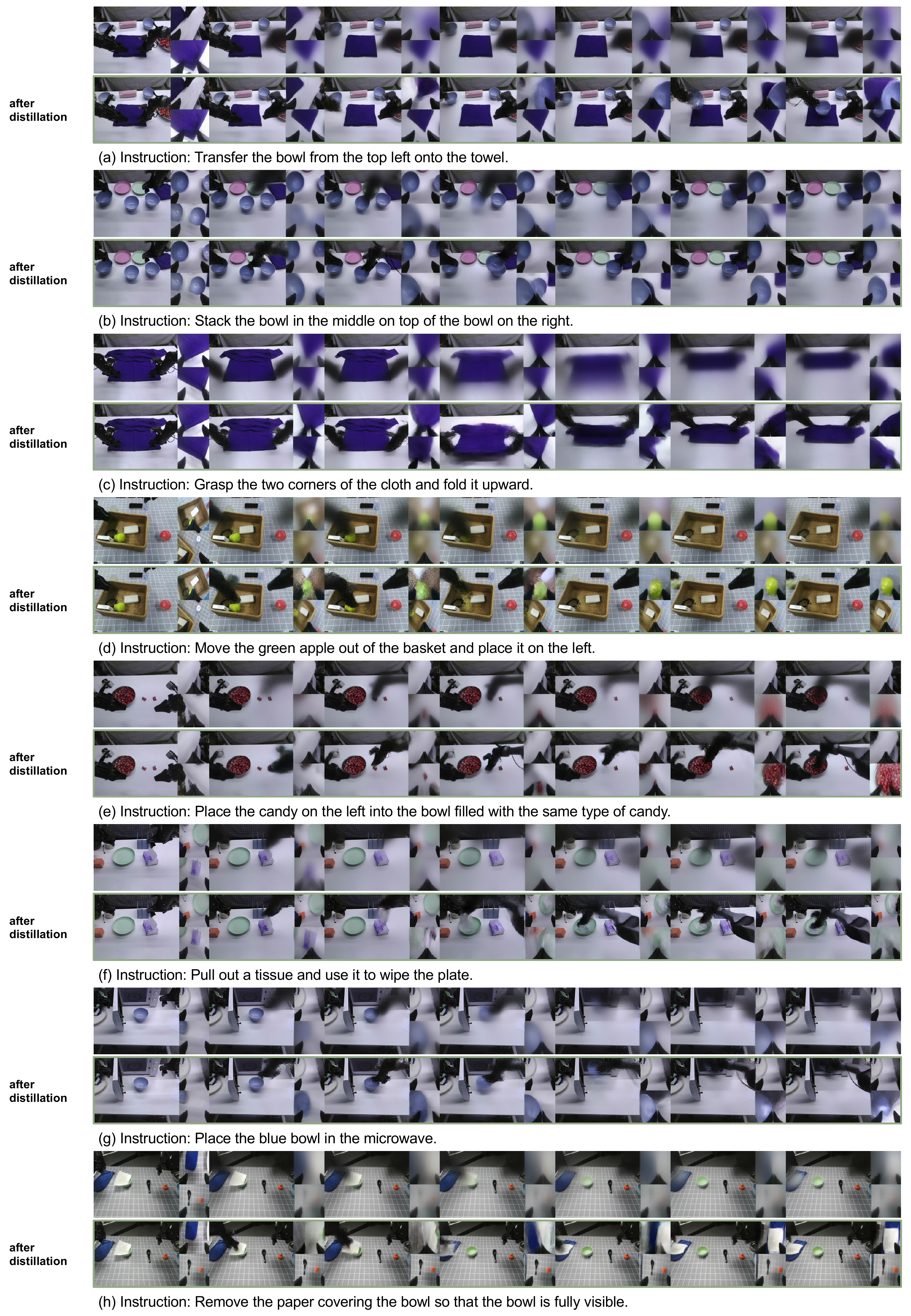}
    \caption{Additional video prediction results on open-ended tasks. We compare \textbf{single-step} video predictions before and after distillation. Distillation enables high-quality video prediction with just one sampling step.}
    \label{fig:video_app}
\end{figure}

\clearpage
\section{Detailed Benchmark Results}\label{app_results}
\begingroup
\setlength{\intextsep}{4pt}
\setlength{\textfloatsep}{6pt}
\setlength{\abovecaptionskip}{3pt}
\renewcommand{\arraystretch}{0.92}
We provide per-suite and per-dimension success rates (\%) for the results in Tables~\ref{tab:libero_generalization} and~\ref{tab:robodojo}, using the same training and evaluation settings as the main text.

\subsection{LIBERO-ID: Per-Suite Success Rates}\label{app_libero}
Table~\ref{tab:app_libero} reports the four standard LIBERO suites~\citep{liu2023libero}. Baseline values follow the original papers, \citet{zhang2026harness}, and \citet{mishra2026understanding}.

\begin{table}[!htbp]
\centering
\small
\begin{tabular}{lccccc}
\toprule
Method & Spatial & Object & Goal & Long & Avg. \\
\midrule
$\pi_0$~\citep{black2024pi_0} & 96.8 & 98.8 & 95.8 & 85.2 & 94.2 \\
$\pi_{0.5}$~\citep{intelligence2025pi_} & 98.8 & 98.2 & 98.0 & 92.4 & 96.9 \\
MolmoAct~\citep{lee2025molmoact} & 87.0 & 95.4 & 87.6 & 77.2 & 86.6 \\
X-VLA~\citep{zheng2026xvla} & 98.2 & 98.6 & 97.8 & 97.6 & 98.1 \\
AtomVLA~\citep{sun2026atomvla} & 96.4 & 99.6 & 97.6 & 94.4 & 97.0 \\
\midrule
Cosmos-Policy~\citep{kim2026cosmospolicy} & 98.1 & \textbf{100.0} & 98.2 & 97.6 & 98.5 \\
Fast-WAM~\citep{yuan2026fastwam} & 98.2 & \textbf{100.0} & 97.0 & 95.2 & 97.6 \\
DiT4DiT~\citep{ma2026dit4dit} & 98.4 & 99.6 & \textbf{98.6} & 97.6 & 98.6 \\
Temporal Ratio~\citep{mishra2026understanding} & 96.3 & 99.6 & 97.6 & 82.6 & 94.0 \\
\midrule
\textbf{VPP2 (Ours)} & \textbf{99.2} & 99.8 & 98.2 & \textbf{98.0} & \textbf{98.8} \\
\bottomrule
\end{tabular}
\caption{LIBERO-ID success rates (\%). Long denotes LIBERO-10. Avg.\ reproduces the reported overall result. Best results in each column are in bold.}
\label{tab:app_libero}
\end{table}

\subsection{LIBERO-Pro: Suite-by-Perturbation Results}\label{app_libero_pro}
Table~\ref{tab:app_libero_pro} follows the eight-column LIBERO-Pro~\citep{zhou2025liberopro} layout of Harness VLA~\citep{zhang2026harness}: \textit{T} denotes instruction redirection and \textit{S} denotes position swap. Available baseline breakdowns are taken from Table~3 of \citet{zhang2026harness}.

\begin{table}[h]
\centering
\small
\setlength{\tabcolsep}{3pt}
\begin{tabular}{@{}lccccccccc@{}}
\toprule
Method & Spat-T & Spat-S & Obj-T & Obj-S & Goal-T & Goal-S & L10-T & L10-S & Overall \\
\midrule
$\pi_0$ & 0.0 & 0.0 & 0.0 & 2.0 & 0.0 & 0.0 & 0.0 & 0.0 & 0.3 \\
$\pi_{0.5}$ & 1.0 & 20.0 & 1.0 & 17.0 & 2.0 & \textbf{38.0} & 1.0 & 8.0 & 11.0 \\
MolmoAct & 0.0 & 0.0 & 0.0 & 6.0 & 0.0 & 0.0 & 6.0 & 0.0 & 1.5 \\
X-VLA & 0.0 & 0.0 & 8.0 & 2.0 & 9.0 & 1.0 & 10.0 & 0.0 & 3.8 \\
AtomVLA & 1.0 & 16.0 & 0.0 & 10.0 & 11.0 & 2.0 & 9.0 & 1.0 & 6.3 \\
\midrule
% VPP2 source: vpp2-policy/runs/libero_policy_eval/horizontal_4suite_harnessvla_public/
% 2026-08-21_bigbs_action30k_public800_local8/comparison_with_rlinf_pi05.json (ours).
% Verified against all 80 task-result JSONs: 100 episodes per column, 360/800 overall.
\textbf{VPP2 (Ours)} & \textbf{58.0} & \textbf{54.0} & \textbf{47.0} & \textbf{58.0} & \textbf{55.0} & 37.0 & \textbf{31.0} & \textbf{20.0} & \textbf{45.0} \\
\bottomrule
\end{tabular}
\caption{LIBERO-Pro success rates (\%) for each suite--perturbation pair. Spat, Obj, Goal, and L10 denote Spatial, Object, Goal, and LIBERO-10. VPP2 uses 100 episodes per column (800 total). Best results in each column are in bold.}
\label{tab:app_libero_pro}
\end{table}

\subsection{LIBERO-OOD: Per-Axis Success Rates}\label{app_libero_ood}
Table~\ref{tab:app_libero_ood} reports the Spatial, Object, and Goal axes of LIBERO-OOD~\citep{li2025vlas}, following \citet{mishra2026understanding}. Neither LIBERO-Pro nor LIBERO-OOD uses additional training on its evaluation tasks; only methods with available results are included.

\begin{table}[!htbp]
\centering
\small
\setlength{\tabcolsep}{8pt}
\begin{tabular}{lcccc}
\toprule
Method & Spatial & Object & Goal & Avg. \\
\midrule
$\pi_0$~\citep{black2024pi_0} & 0.7 & 0.3 & 4.3 & 1.7 \\
$\pi_{0.5}$~\citep{intelligence2025pi_} & 36.7 & 2.3 & 41.7 & 26.8 \\
\midrule
Cosmos-Policy~\citep{kim2026cosmospolicy} & 30.7 & 0.3 & 0.7 & 10.5 \\
Fast-WAM~\citep{yuan2026fastwam} & 12.7 & 0.0 & 15.7 & 9.4 \\
DiT4DiT~\citep{ma2026dit4dit} & 9.0 & 0.0 & 10.3 & 6.4 \\
Temporal Ratio~\citep{mishra2026understanding} & 58.6 & 40.0 & 80.3 & 59.4 \\
\midrule
\textbf{VPP2 (Ours)} & \textbf{59.3} & \textbf{44.5} & \textbf{87.7} & \textbf{63.9} \\
\bottomrule
\end{tabular}
\caption{LIBERO-OOD success rates (\%) for Spatial, Object, and Goal. Avg.\ is the reported overall success rate; best results in each column are in bold.}
\label{tab:app_libero_ood}
\end{table}

\clearpage
\endgroup
\subsection{RoboDojo: Per-Dimension Success Rates and Scores}\label{app_robodojo}
RoboDojo~\citep{chen2026robodojo} evaluates generalist manipulation policies on 42 bimanual simulation tasks organized into five capability dimensions. \textit{Generalization} tests robustness to unseen backgrounds, lighting, clutter, and target objects, and is evaluated under both standard and randomized settings. \textit{Precision} requires fine-grained target localization and contact-rich control. \textit{Long-Horizon} requires completing all sub-steps of multi-step tasks. \textit{Memory} contains tasks whose correct actions depend on information observed earlier in the episode. \textit{Open} evaluates unseen task specifications whose required skills appear in the training data under different contexts. We follow the official training and evaluation protocol. The average score captures partial task progress, and the success rate measures binary task completion.

Table~\ref{tab:app_robodojo_sr} reports success rates for each capability dimension, and Table~\ref{tab:app_robodojo_score} provides the corresponding average scores. These results complement Table~\ref{tab:robodojo}. VPP2 uses full-task instructions without subtask planning. Results are taken from the \href{https://robodojo-benchmark.com/leaderboard}{official RoboDojo simulation leaderboard}, accessed on October 12, 2026. Generalization results are reported separately for standard (Std.) and randomized (Rand.) settings for all methods, including VPP2.

\begin{table}[!htbp]
\centering
\small
\setlength{\tabcolsep}{3.5pt}
\begin{tabular}{lccccccc}
\toprule
 & \multicolumn{2}{c}{Generalization} & & & & & \\
\cmidrule(lr){2-3}
Method & Std. & Rand. & Precision & Long-Horizon & Memory & Open & Avg. \\
\midrule
$\pi_0$~\citep{black2024pi_0} & 4.89 & 0.22 & 0.75 & 2.00 & 2.11 & 0.25 & 1.53 \\
$\pi_{0.5}$~\citep{intelligence2025pi_} & 14.89 & 1.44 & 5.50 & 14.67 & 4.67 & 1.67 & 6.93 \\
X-VLA~\citep{zheng2026xvla} & 12.22 & 1.33 & 12.00 & 9.75 & 3.56 & 0.50 & 6.52 \\
Fast-WAM~\citep{yuan2026fastwam} & 2.11 & 0.11 & 0.00 & 5.17 & 3.44 & 0.42 & 2.03 \\
OpenWAM-$\alpha$~\citep{wang2026openwam} & 25.56 & 4.11 & 9.25 & 25.33 & 9.11 & 1.08 & 11.92 \\
Xiaomi-Robotics-1~\citep{guo2026xiaomirobotics1} & 28.00 & 6.00 & 18.83 & 23.67 & 6.56 & 3.58 & 13.93 \\
GPT-6-Astra~\citep{zhang2026astra} & 32.67 & 28.33 & 4.00 & 8.25 & 38.67 & 31.00 & 22.48 \\
\midrule
% Official leaderboard: generalizationStd successRate=46.44; generalizationRand successRate=12.56.
\textbf{VPP2 (Ours)} & 46.44 & 12.56 & 35.17 & 37.00 & 55.56 & 4.08 & \textbf{32.26} \\
\bottomrule
\end{tabular}
\caption{Per-dimension success rates (\%) on the RoboDojo simulation benchmark. Avg.\ equally weights the five capability dimensions, with Generalization computed as the mean of Std.\ and Rand.}
\label{tab:app_robodojo_sr}
\end{table}

\begin{table}[!htbp]
\centering
\small
\setlength{\tabcolsep}{3.5pt}
\begin{tabular}{lccccccc}
\toprule
 & \multicolumn{2}{c}{Generalization} & & & & & \\
\cmidrule(lr){2-3}
Method & Std. & Rand. & Precision & Long-Horizon & Memory & Open & Avg. \\
\midrule
$\pi_0$~\citep{black2024pi_0} & 7.18 & 0.71 & 3.56 & 6.19 & 3.47 & 0.25 & 3.48 \\
$\pi_{0.5}$~\citep{intelligence2025pi_} & 20.93 & 5.82 & 12.40 & 23.54 & 5.89 & 1.98 & 11.44 \\
X-VLA~\citep{zheng2026xvla} & 17.90 & 3.04 & 18.32 & 16.53 & 4.76 & 0.55 & 10.13 \\
Fast-WAM~\citep{yuan2026fastwam} & 4.33 & 0.34 & 1.96 & 9.14 & 3.55 & 0.42 & 3.48 \\
OpenWAM-$\alpha$~\citep{wang2026openwam} & 33.16 & 8.26 & 18.45 & 34.93 & 10.41 & 1.41 & 17.18 \\
Xiaomi-Robotics-1~\citep{guo2026xiaomirobotics1} & 35.65 & 11.44 & 26.69 & 38.39 & 7.81 & 3.94 & 20.07 \\
GPT-6-Astra~\citep{zhang2026astra} & 35.32 & 31.40 & 12.65 & 21.45 & 43.04 & 34.36 & 28.97 \\
\midrule
% Official leaderboard: generalizationStd score=52.14; generalizationRand score=20.63.
\textbf{VPP2 (Ours)} & 52.14 & 20.63 & 43.03 & 56.28 & 56.01 & 4.61 & \textbf{39.26} \\
\bottomrule
\end{tabular}
\caption{Per-dimension average scores on the RoboDojo simulation benchmark. Scores capture partial task progress and use the same five-dimension aggregation as the success rates in Table~\ref{tab:app_robodojo_sr}.}
\label{tab:app_robodojo_score}
\end{table}

\end{document}